\documentclass{llncs}

\usepackage[]{algorithm2e}
\usepackage{algorithmic}
\usepackage{graphicx, subfigure}
\usepackage{color}
\usepackage{float}
\usepackage{amsmath}
\usepackage{soul}
\usepackage{tabu}

\DeclareMathOperator*{\argminG}{arg\,min}

\begin{document}

\title{ORBSLAM-based Endoscope Tracking and 3D Reconstruction}
%
%\titlerunning{Hamiltonian Mechanics}  % abbreviated title (for running head)
\author{Nader Mahmoud \inst{1,2} \and I{\~n}igo Cirauqui \inst{3} \and Alexandre Hostettler \inst{1} \and Christophe Doignon \inst{2} \and Luc Soler \inst{1} \and Jacques Marescaux \inst{1} \and J.M.M. Montiel \inst{3}}
\authorrunning{gggggggggg} % abbreviated author list (for running head)
\tocauthor{*********}

\institute{
IRCAD (Institut de Recherche contre les Cancers de l'Appareil Digestif), France,\\
\email{nader-mahmoud.ali@etu.unistra.fr}
\and
ICube (UMR 7357 CNRS), Universit{\'e} de Strasbourg, France\\
\and
Instituto de Investigaci{\'o}n en Ingenier{\'i}a de Arag{\'o}n (I3A), Universidad de Zaragoza, Spain,
\email{josemari@unizar.es}}

\maketitle              % typeset the title of the contribution

\begin{abstract}
We aim to track the endoscope location inside the surgical scene and provide 3D reconstruction, in real-time, from the sole input of the image sequence captured by the monocular endoscope. This information offers new possibilities for developing surgical navigation and augmented reality applications. The main benefit of this approach is the lack of extra tracking elements which can disturb the surgeon performance in the clinical routine. It is our first contribution to exploit ORBSLAM, one of the best performing monocular SLAM algorithms, to estimate both of the endoscope location, and 3D structure of the surgical scene. However, the reconstructed 3D map poorly describe textureless soft organ surfaces such as liver. It is our second contribution to extend ORBSLAM to be able to reconstruct a semi-dense map of soft organs. Experimental results on in-vivo pigs, shows a robust endoscope tracking even with organs deformations and partial instrument occlusions. It also shows the reconstruction density, and accuracy against ground truth surface obtained from CT.

\end{abstract}
\section{Introduction}
Minimally Invasive Surgery (MIS) practice has several drawbacks for the surgeon, such as, lack of depth perception, or poor localization within operating field due to the limited field of view. The intra-operative 3D reconstruction of surgical scene simultaneous to tracking endoscope position in real-time provides key information for many MIS tasks. These tasks include surgical navigation (in case of flexible endoscope), and Augmented Reality (AR) overlies of pre-operative medical data in the endoscope live video stream. 
% However, 3D scene reconstruction from endoscope images of soft organs remains a difficult problem. Due to organs deformations, instrument occlusion and lack of good features to be tracked. 

Recently, computer vision based algorithms have attracted the attention, for their success in  providing intra-operative reconstruction of the surgical scene, and the tracking of the stereo-endoscope position \cite{Stoyanov,Lin}. However, these methods are not adapted to the commonly used monocular endoscope. Structure from motion (SfM) methods have been proposed to deal with monocular endoscope \cite{Wu,Sun}. However, SFM methods requires offline batch processing, what makes them not suitable for real-time applications. Therefore, in \cite{Sun} a tracking sensor is attached to the endoscope to estimate its position. 

VSLAM (Simultaneous Location And Mapping from Visual sensor) is a popular topic in robotics, which aims at simultaneously building a 3D map of unknown environment while keep track of camera location. VSLAM use in MIS has been researched by Mountney et al. \cite{Mountney}, who applied and extended the Extended Kalman Filter SLAM (EKF-SLAM) framework from Davison \cite{Davison2003} to MIS environment, but with stereo-endoscope. For periodic liver deformation, Mountney and Yang \cite{MountneyMotion} proposed to learn the parameters of the periodic motion first, and then use it to improve the VSLAM estimation. 

In \cite{Klein}, Klein and Murray proposed the Parallel Tracking and Mapping (PTAM)  algorithm that represented a breakthrough in visual SLAM. Lin et al. \cite{Lin} adapted PTAM to a stereo-endoscope in order to reconstruct a denser 3D map than those made by EKF-SLAM systems. Due to non-rigid deformation in surgical scenes, the use of only a monocular endoscope has proven challenging. Grasa et.al. \cite{Grasa} provided experimental evidence of the feasibility of monocular EKF SLAM in medical scenes. In \cite{Grasa14}, they provided extensive validation on in-vivo human sequences proofing its ability to be used for hernia defect measurements in hernia repair surgery.

Following the venue open by PTAM,  the ORBSLAM system \cite{Mur-Artal} has been proposed recently, it has proven as a robust camera tracking and mapping estimator with remarkable camera relocation capabilities. Our first contribution is researching ORBSLAM performance within MIS environments. By only re-tuning the system, the endoscope location was robustly tracked and relocated successfully after tracking loss. However, it is at the expense of a low map density, mainly due to the lack of repeatability of the ORB features in some body structures such as the liver. It is also our contribution a new matching algorithm to densify the map and hence improve the estimated 3D map.
In the experimental, section we provide qualitative evaluation of the performance in several in-vivo pig sequences, including respiration, and tools cluttering the endoscope field of view. We also provide a quantitative assessment that yields an accuracy in the range between 3mm to 4.5mm when the VSLAM map points are compared with respect to a ground truth surface from CT. Additionally, the tracked endoscope location has been exploited to provide support for augmented reality overlays of preoperative models onto the endoscope live video stream.

\section{ORBSLAM overview}
ORBSLAM is based on keyframes and nonlinear optimization as proposed in PTAM. It includes the covisibility information in the form of a graph as proposed in \cite{Strasdat}, in addition to bag of binary words DBoW2 proposed in \cite{GalvezLopez} for place recognition. For large scale mapping, scale aware loop closing \cite{StrasdatScaleAware} is used. The system uses ORB \cite{Rublee} for feature detection and description in all processes, what boots the performance in the place recognition and loop closure operations. A complete description of the algorithm can be found in \cite{Mur-Artal}. For the sake of completeness, we summarize next the more relevant steps: tracking, mapping and relocation.

\paragraph{Tracking}
This task tracks the endoscope location sequentially in every frame of the live video. The 3D locations of the map points are assumed to be available, each of them with a valid ORB binary descriptor. At the current frame, an initial guess for the endoscope position is estimated from the previous frame by means of a motion model, then the map points are reprojected to estimate its image in the current frame. The ORB descriptor of each map point is compared with those of all the features detected inside a search region surrounding the predicted point. The feature point in the image with the smallest Hamming distance is selected as the match, only if it is over a threshold. Then the pose of the frame is refined by Huber robustified non-linear optimization of the reprojection error for the matched points. After the optimization stage, the matches are segmented as inliers or outliers according to the Huber threshold. Map points rendering outlier matches consistently during initialization are considered non reliable and do not survive in the initialization process.

\paragraph{Mapping.}
To build the 3D map of the scene, the system selects a set of frames from the endoscope sequence. This selected frames are called keyframes. Benefiting from the matches provided by the tracking process, the system estimates matches across the keyframes. Once the matches are available, the 3D location for the map points and the 3D poses for the keyframes are computed by bundle adjustment (BA). The algorithm sequentially computes the matches and iteratively improves the map accuracy, in a thread that runs in parallel with the tracking thread, but at lower frequency. The BA minimizes total Huber robustified reprojection error with respect to the keyframe positions, $\textbf{X}_{WC_i}$, and the 3D map point locations, $\textbf{X}_{Wj}$:
\begin{equation}
\argminG_{\scriptsize \textbf{X}_{Wj}, \textbf{X}_{WC_i}}\sum_{i,j}{\rho\left(\left||\textbf{u}_{ij} - \mbox{CamProj}(\textbf{X}_{Wj},\textbf{X}_{WC_i})\right||\right)}
\end{equation}
where $\textbf{u}_{ij}$ is the matched observation of the \textit{j}-th map point by the \textit{i}-th keyframe. \mbox{CamProj} codes the projection function including perspective and radial distortion. $\rho$ denotes the robust Huber influence function. As the endoscope explores new areas of the scene not imaged previously, new keyframes are added to the map. After adding a new keyframe, new matches with respect to the previous keyframes are found to initialize new map points.

Initially, map points and keyframes are initialized in excess, then in a second stage a demanding rigorous is applied to select the fittest to survive. The reasons for culling a map point are: (1) The point cannot be tracked and matched in the following frames. (2) The projection rays used to triangulate the point in 3D render low parallax. (3) The triangulated point produces excessive reprojection error over the keyframes where it is observed. This severe selection of points have proven essential for robust performance in endoscope sequences. The keyframes whose 90\% of the map points have been detected in at least other three keyframes are removed from the map, in order to keep just the more informative ones.

Map points are initialized by detecting ORB features at different image scales to achieve both scale and rotation invariance. One of the strong points of the algorithm is that ORB features are used both for mapping, and for the place recognition. Place recognition combines a Bag of Words built from the ORB binary descriptors, with the covisibility graph that determines all the keyframes that are observing the same 3D scene region. 

\paragraph{Endoscope relocation.}
Tacking can be lost because of occlusion, feature deletion due to fast endoscope motion, or failure to match enough map points. Therefore, the endoscope has to be located with respect to the map from scratch. Relocation is also known as the kidnapped camera situation. All the keyframes of the map are stored in a Bag of Binary Words indexed database to recover the more similar keyframes in response to a query image. More crucially, thanks to the covisibility graph, the set of keyframes observing the same area of the map can be also recovered. After tracking loss, the ORB detected in the image gathered by the endoscope are used to query the database to detect the set of keyframes that are observing the same scene area as the endoscope image. Additionally, the system also provides a set of putative ORB matches between the image and a set of 3D points in the map. Then endoscope position is estimated by P3P and RANSAC. Once a valid endoscope pose is estimated, the tracking can be resumed.

\section{Extending the map density}
The mapping thread is responsible for creation/deletion of map points, and map refinement through BA. After new keyframe arrival, all of its ORB features are matched against closest keyframes, and all matched ORB points are triangulated and appended to the map. However, map points cannot be initialized on soft organs like liver, because they can not be repetitively detected along several frames in the sequence. We extended this initialization process to a second stage. Firstly, all matched ORB points are triangulated. Secondly, a cross-correlation guided by epipolar geometry is used to find matches for all unmatched ORB points in the newly added keyframe, according to Algorithm 1. 

\begin{figure}[H]
\centering
  \subfigure[]{
    \includegraphics[width=0.25\textwidth]{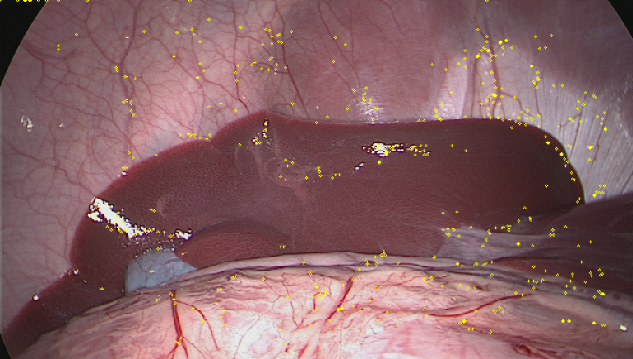}
   }
  \subfigure[]{
    \includegraphics[width=0.25\textwidth]{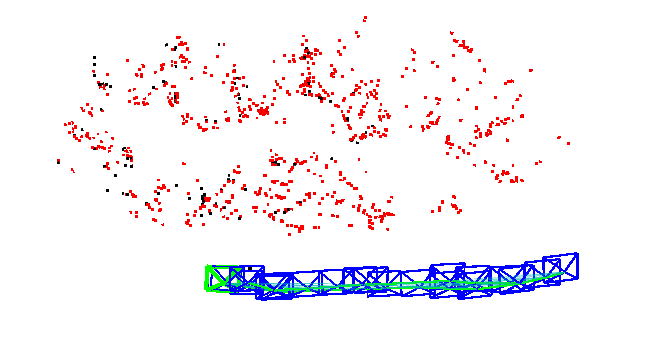}
   }
   \\
   \subfigure[]{
    \includegraphics[width=0.25\textwidth]{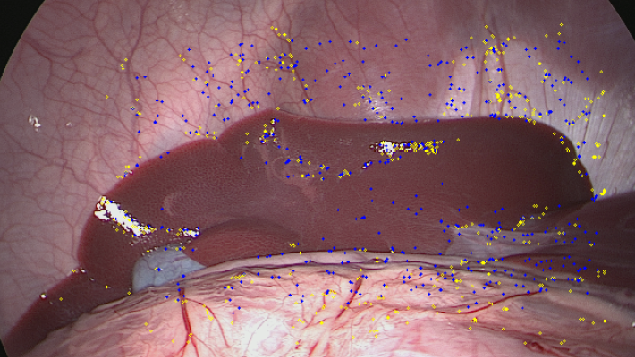}
   }
   \subfigure[]{
    \includegraphics[width=0.25\textwidth]{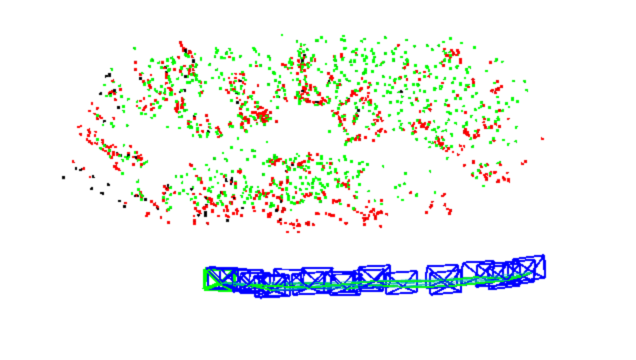}
   }
   \subfigure[]{
    \includegraphics[width=0.25\textwidth]{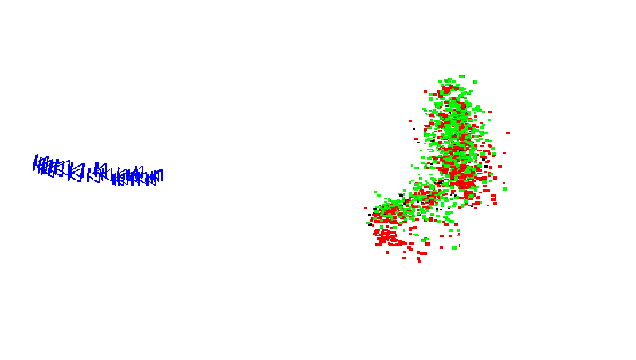}
   }
\caption{Algorithm 1 semi-dense map. (a) Reprojection of the ORB points (yellow). (b) ORBSLAM 3D map. (c) Reprojection of algorithm 1 map points. (d, e) Algorithm 1 map (green) and ORBSLAM map (red), from two different points of view}
\label{fig:result of algorithm1}
\end{figure}

Fig. \ref{fig:result of algorithm1} shows the map obtained by Algorithm 1. The original map created by ORBSLAM and its reprojection onto one liver image are shown in Fig. \ref{fig:result of algorithm1}(a). Blue  rectangles in Fig. \ref{fig:result of algorithm1}(b,d,e) are keyframes describing endoscope trajectory,  camera position in current image is displayed in green. Red points are ORB map points. A semi-dense map is obtained by reconstructing points in a sparse regions in the image and represented as green points in Fig. \ref{fig:result of algorithm1}(d,e). More points will be reconstructed when exploring new regions. In subsequent frames, the newly reconstructed points (by algorithm 1) are tracked, firstly, by Lucas-Kanade optical flow. Secondly, for all untracked points a cross-correlation search guided by epipolar geometry is performed using patch around the point  extracted in the keyframe used for its 3D triangulation in Algorithm 1.

%\vspace{-2mm}
\begin{algorithm} [H]
	\SetAlgoLined
	\SetKwData{Left}{left}\SetKwData{This}{this}\SetKwData{Up}{up}
    \SetKwFunction{Union}{Union}\SetKwFunction{FindCompress}{FindCompress}
    \SetKwInOut{Input}{Input}
    \SetKwInOut{Output}{Output}
	\ForEach{Newly added KF ($KF_C$)}{
    	 Get 4 neighbors KFs with significant baseline\\
         \ForEach{Neighbor KF ($KF_N$)}{         
           \ForEach{unmatched ORB feature ($P_c$) in $KF_C$}{
           		- Extract a rectangular correlation patch\\           	
              	- Patch crosscorrelation around epipolar segment in $KF_N$\\
              	- Threshold on maximal distance to epipolar line\\
              	- Triangulate map point from matched two observations\\
              	- Remove points with negative depth relative to $KF_C$ and $KF_N$\\
              	- Threshold on maximal reprojection error onto $KF_C$ and $KF_N$ \\ 
                - Remove point if depth different from median depth in $KF_C$\\       
		  }
     	}
    }
    \caption{Cross-correlation search for 3D point triangulation}
\end{algorithm}

\section{Experimental results}
The performance was evaluated with several in-vivo pig laparoscopy sequences. The endoscope camera was calibrated following \cite{Zhang}. Next we detail the different experiments. More details can be appreciated in our video \cite{video}
\subsection{ORBSLAM performance evaluation}
We re-tune the ORBSLAM to overcame the  key factors limiting its performance when processing endoscope sequences, we report modifications relative to the ORBSLAM standard rigid case:
\begin{description}
\item [Search region in tracking].- For tracking, map points are reprojected in the image, and each one of them defines a search region in which a match with an image keypoint is attempted. We have increased the size of the search region in a factor 1.5 (0.5 pixel), not to loose some matches due to potential deformation.
\item [Parallax threshold at point initialization].- When a map point is created, it is enforced to have at least a threshold parallax to ensure that its location in 3D is accurate. Minimum parallax is increased in a factor of 5, it becomes 1.4035$^{\circ}$, to increase the accuracy in the triangulated points.
\item[Reprojection error threshold].- A maximum threshold is allowed in the distance between the reprojected map point and the image keypoints used for its triangulation. We reduced this threshold in a factor 10, it becomes 0.5991, to ensure that only rigid scene points are included eventually in the map.
\item[Hamming distance threshold].- We reduce the allowed Hamming distance between descriptors of matched image points. We decreased it by a factor 0.9, it becomes 45 bit, to enforce more similarity in the accepted matches.
\end{description}

\begin{figure}
\centering
  \subfigure[]{
  	\label{perf1a}
    \includegraphics[height=0.18\textwidth]{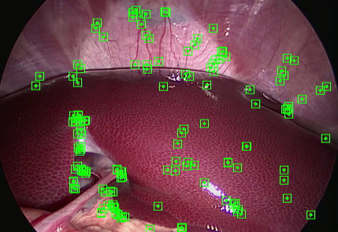}}
   \subfigure[]{
   	\label{perf1b}
    \includegraphics[height=0.18\textwidth]{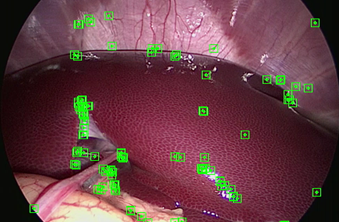}}
   \subfigure[]{
   	\label{perf1c}
    \includegraphics[height=0.18\textwidth]{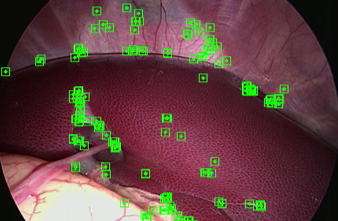}}
    
   \subfigure[]{
   	\label{perf1d}
    \includegraphics[width=0.21\textwidth]{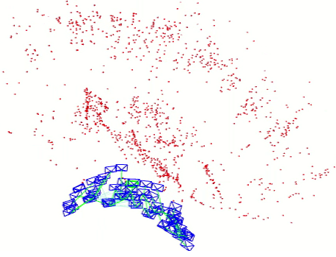}}
   \subfigure[]{
   	\label{perf1e}
    \includegraphics[width=0.21\textwidth]{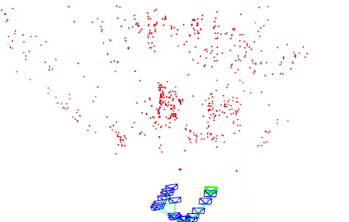}}
   \subfigure[]{
   	\label{perf1f}
    \includegraphics[width=0.21\textwidth]{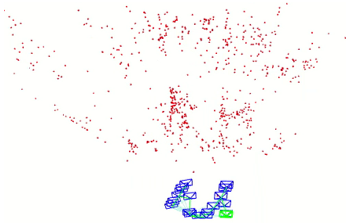}}
    
\caption{ORBSLAM performance. (a-c) images with reprojected map points(green points). (d) Reconstructed map points (red) and keyframes (endoscope tip trajectory). (e-f) Breathing motion, current endoscope location is shown as a green rectangle. (e) and (f) during inhale and exhale, respectively.}
\label{fig:PerformanceEval}
\end{figure}

We have found that the endoscope tracking qualitatively quite robust and accurate. However, there are many areas of the scene where the system is unable to track map points, being able to match only 24\% of the map points visible in the image. The main reason for this failure in matching, around 50\% of the potential matches, is that ORB detector is not able to detect repeatable points on soft organs, such as liver. Also, BA in mapping process considers 11\% of the map points as non-rigid, this percentage raises up to 25\% in areas with visually high non-rigid component. Despite the low number of matched map points, the system was able to compute an sparse map. Fig. \ref{perf1d} shows the reconstructed map which consisted of 66 keyframes and 1566 map points. In this part of the sequence the endoscope was fixed relative to  the operating table, Fig . \ref{perf1e} and \ref{perf1f} show the ability to estimate the breathing motion, because of the  pig breathing there was a forward-backward motion of the diaphragm able to be seen in the camera trajectory.

\begin{figure} [H]
\centering
  \subfigure[]{
  	\label{reloc2a}
    \includegraphics[width=0.9\textwidth]{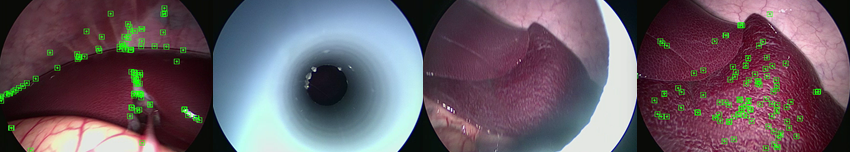}}
   \subfigure[]{
   	\label{reloc2b}
    \includegraphics[trim={0 0.4cm 0 1cm},clip,width=0.4\textwidth]{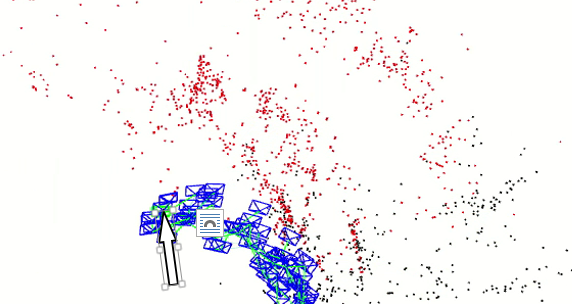}}
   \subfigure[]{
   	\label{reloc2c}
    \includegraphics[trim={0 0.4cm 0 1cm},clip,width=0.4\textwidth]{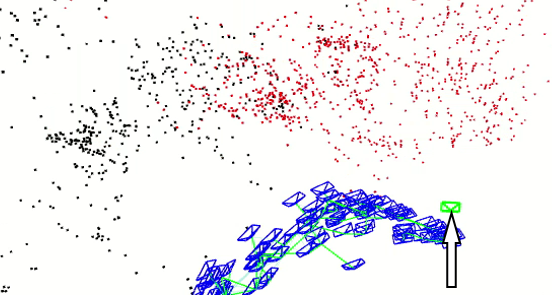}}
\caption{Relocation ability. (a) Consecutive stages from left to right: successful tracking while observing the liver, tracking loss when endoscope was extracted, endoscope inserted again imaging the spleen, relocation. (b,c) The arrows refer to the  endoscope locations before tracking lost, and after relocation}
\label{fig:Relocation}
\end{figure}

Additionally, the system was able to accurately relocate the endoscope location after tracking loss. In Fig. \ref{fig:Relocation}, after the exploration phase of the abdominal cavity, the endoscope was extracted outside the cavity while looking at the liver, and it is later reinserted imaging the spleen. Since, several spleen points had been  mapped before, the system was able to relocate the endoscope location within 3 seconds.

\begin{figure}[H]
\centering
  \subfigure[]{
  	\label{reloc3a}
    \includegraphics[trim={0 0 0 12mm},clip,height=0.225\textwidth]{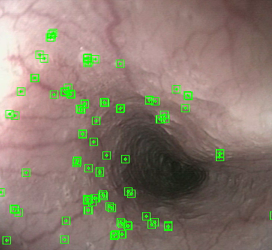}}
   \subfigure[]{
   	\label{reloc3b}
    \includegraphics[height=0.225\textwidth]{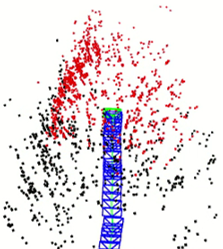}}
\caption{Gastroscopy sequence. (a) Esophagus with tracked points. (b) Reconstructed map (red and black points), keyframes (blue rectangles) and current endoscope location (green rectangle).}
\label{fig:Gastroscopy}
\end{figure}

Finally, the system has also been tested with challenging gastroscopy sequences which contains reflection and abrupt movements. It was able to track the endoscope location and reconstruct 3D map of the scene (cf. Fig. \ref{fig:Gastroscopy}). The average tracking time per frame was approximately 25 ms on desktop PC with Intel(R) Core d
-3337U CPU @ 1.80GHz with 6 GB RAM.

\subsection{Estimation of reconstruction error}
\label{Estimation_of_reconstruction_error}
\begin{figure}
\centering
  \subfigure[]{
    \includegraphics[width=0.25\textwidth]{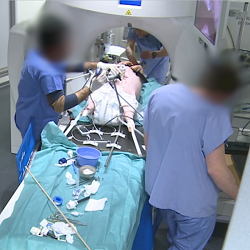}
   } 
   \subfigure[]{
    \includegraphics[width=0.25\textwidth]{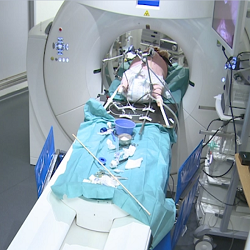}
   }
\caption{Data acquisition. (a) Video recording. (b) CT acquisition while endoscope was fixed and its tip inside the abdominal cavity}
\label{fig:CT_acquisition}
\end{figure}

To evaluate the error associated with the reconstructed point cloud of the scene, two pigs were used inside computed tomography (CT) room to obtain in-vivo sequences with CT ground truth surface.  A monocular endoscope explores the abdominal cavity before any interaction with the liver. Then a CT scan was  performed while the endoscope was fixed by means of an articulated arm as shown in Fig. \ref{fig:CT_acquisition}. In all CT acquisitions, the tip of the endoscope was included in the CT, to be segmented and extracted from the CT images. The length of the recorded sequences ranged between 2 to 10 minutes.
\vspace{-1mm}
\vspace{-4mm}

The abdominal surface was segmented from CT images and considered as ground truth. In order to compare with the the VSLAM map, firstly, the endoscope was segmented from CT images and its position w.r.t the surface was computed using \cite{Sylvain} . Endoscope position estimated by ORBSLAM and by \cite{Sylvain} were aligned, however VLSAM cannot recover the scene scale ($\lambda$), and \cite{Sylvain} cannot recover the endoscope roll angle ($\theta$) so additional scale and rotational alignment was needed before comparing the two scene maps. The alignment is not a critical process, brute-force search to find both the scale and the roll angle that minimize the distance between the VSLAM map and the CT surface was used.

\begin{figure}
\centering
  \subfigure[]{
    \includegraphics[width=0.30\textwidth]{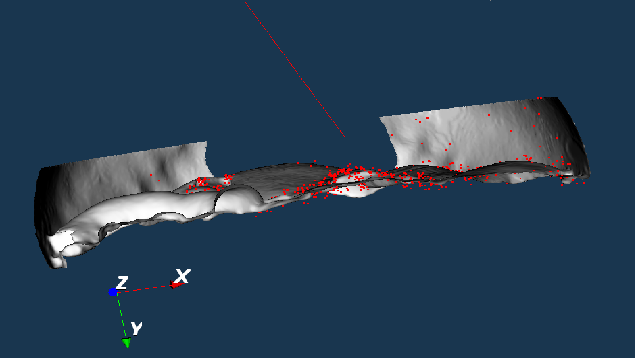}
   } 
   \subfigure[]{
    \includegraphics[width=0.30\textwidth]{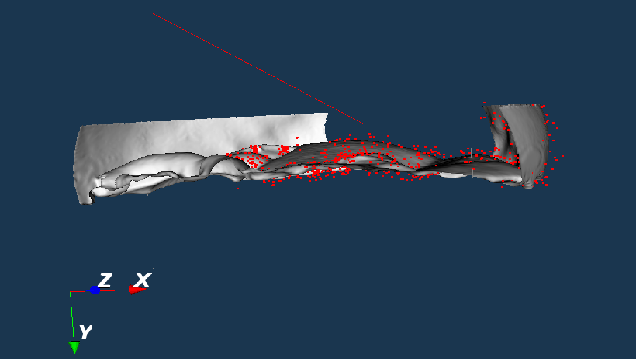}
   } 
   \subfigure[]{
    \includegraphics[width=0.30\textwidth]{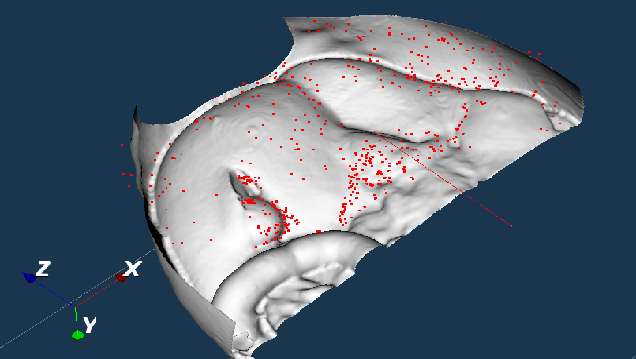}
   } 
\caption{Alignment of point cloud with ground truth surface. (a) Reconstructed ORB points. (b) and (c) alignment of ORB points and points obtained by algorithm 1 from different directions. Red line is the estimated endoscope position by \cite{Sylvain}.}
\label{fig:pointcloud2surface}
\end{figure}

The distance is defined as the euclidean distance between each point in the VSLAM map its closest one on the CT surface. The closest point on the surface is the one with smallest perpendicular distance. So the the cost function for the Brute-force optimization is:
\begin{equation} 
\argminG_{\scriptsize \lambda,R(\theta)} \sqrt{\frac{1}{N}\sum_{i=1}^{N} ||P_i - \lambda \cdot R(\theta) \cdot Q_i||^2}
\label{eq1}
\end{equation}
where $P_i$ are CT surface points closest point and $Q_i$ are the $N$ VSLAM map points. $\lambda$ and $R(\theta)$ are the scale factor and rotation matrix calculated from the roll angle, respectively. Only 80\% of the points are considered in RMSE computation. The remaining 20\% are either outliers or points reconstructed on the diaphragm wall which was outside the CT field of acquisition. The obtained RMSE of considering only reconstructed ORB points was approximately 3 mm (cf. Fig. \ref{fig:pointcloud2surface}(a)). The RMSE  of the semi-dense map obtained by algorithm 1 was approximately 4.1 mm (cf. Fig. \ref{fig:pointcloud2surface}(b-c)).
 
\vspace{-3mm}
\vspace{-8mm}

\subsection{Evaluation with instrument occlusion and deformations}

\begin{figure}
\centering
  \subfigure[]{
    \includegraphics[width=0.22\textwidth]{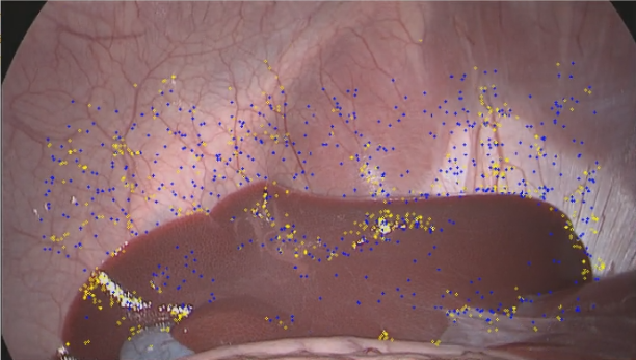}
   }
%   \subfigure[]{
%     \includegraphics[width=0.35\textwidth]{exp1_img_2.png}
%    }  
   \subfigure[]{
    \includegraphics[width=0.22\textwidth]{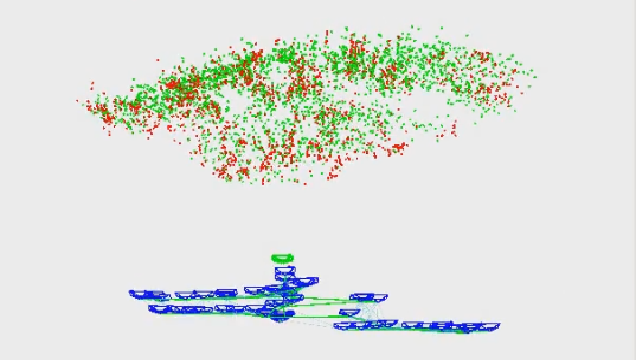}
   }
   \subfigure[]{
    \includegraphics[width=0.22\textwidth]{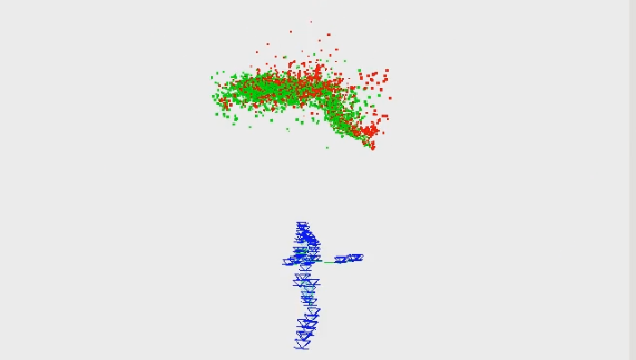}
   }
   \subfigure[]{
    \includegraphics[width=0.22\textwidth]{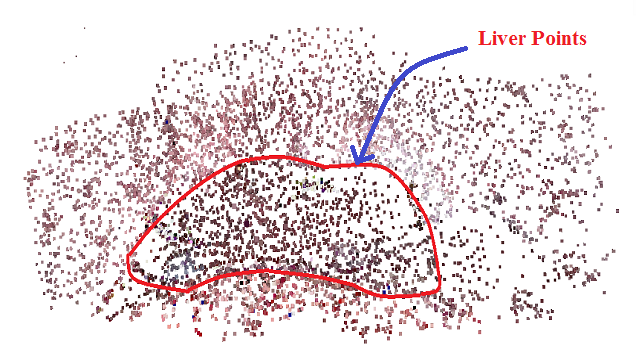}
   }
   
\caption{Endoscope tracking and reconstructed 3D map from exploration phase. (a-c) Reconstructed points and keyframe positions from different directions. (e) Reconstructed points colored using the same RGB color of the 2D features}
\label{fig:result1}
\end{figure}

Several pig liver sequences are used, which contains instruments interacting with the liver, what generates deformations and occlusions. Fig. \ref{fig:result1}(b-d) shows the endoscope tracking on one liver sequence, where red points are reconstructed ORB points and green points are reconstructed by algorithm 1. The estimated endoscope position in the current frame is represented in green rectangle, while the blue rectangles represent the trajectory described as  past keyframe positions. Yellow and blue points in Fig. \ref{fig:result1}(a) are the reprojection of ORB points and points reconstructed by algorithm 1, respectively. We use the same colors for all subsequent figures. As it can be noticed, more points were reconstructed particularly on the liver (Fig. \ref{fig:result1}(e)). The number of the recovered 3D map points were about 4599 with 58 keyframes. 

% \vspace{-1mm}

\begin{figure}
\centering
  \subfigure[]{
    \includegraphics[width=0.25\textwidth]{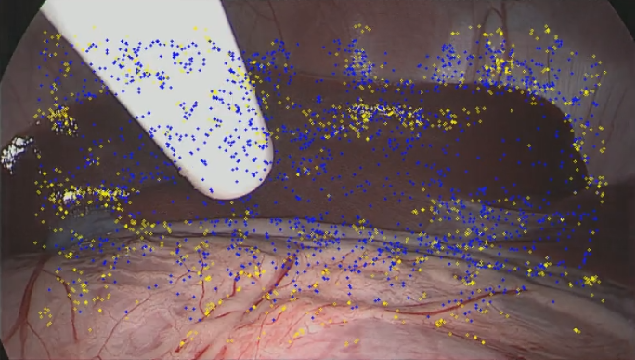}
   }
  \subfigure[]{
    \includegraphics[width=0.25\textwidth]{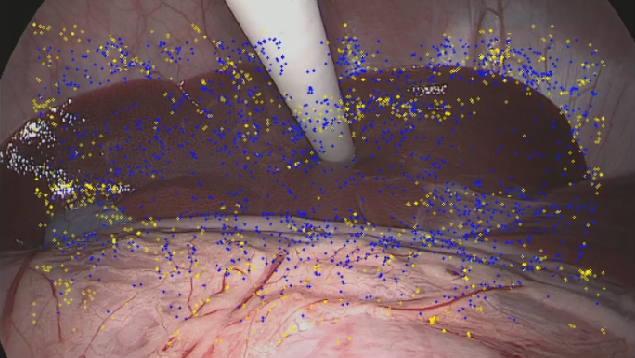}
   }
   \subfigure[]{
    \includegraphics[width=0.25\textwidth]{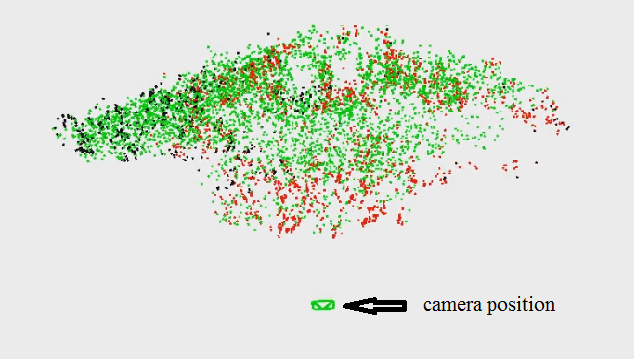}
   }
   \subfigure[]{
    \includegraphics[width=0.25\textwidth]{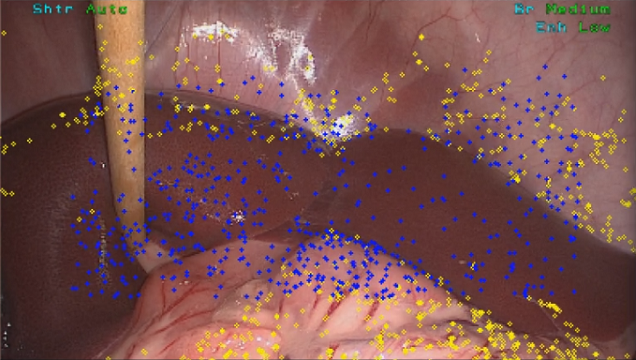}
   }
  \subfigure[]{
    \includegraphics[width=0.25\textwidth]{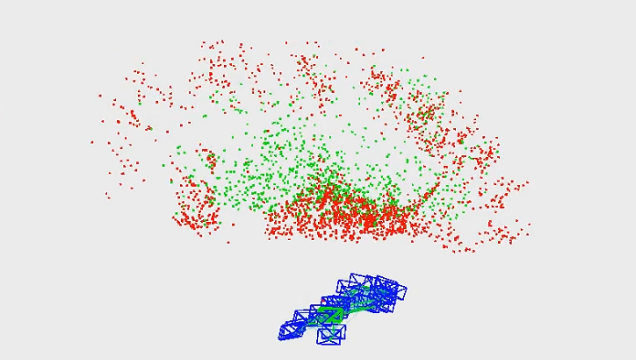}
   }
   \subfigure[]{
    \includegraphics[width=0.25\textwidth]{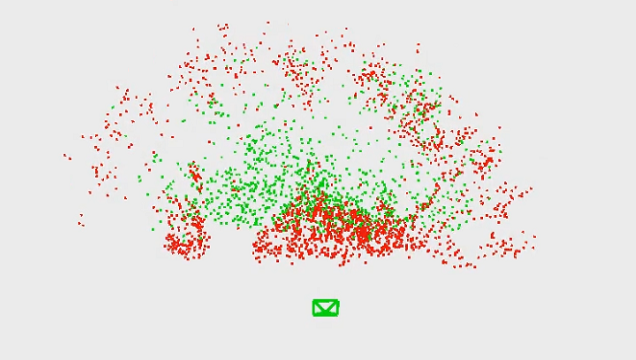}
   }
    \subfigure[]{
    \includegraphics[width=0.25\textwidth]{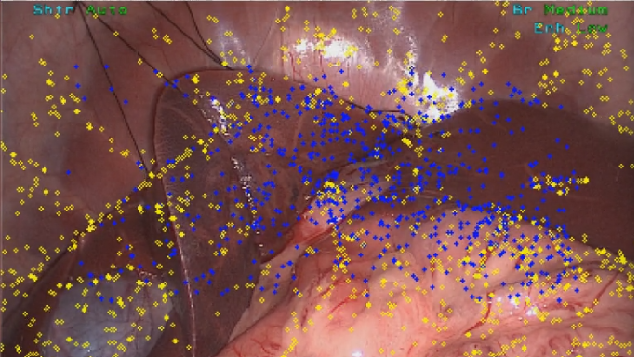}
   }
    \subfigure[]{
    \includegraphics[width=0.25\textwidth]{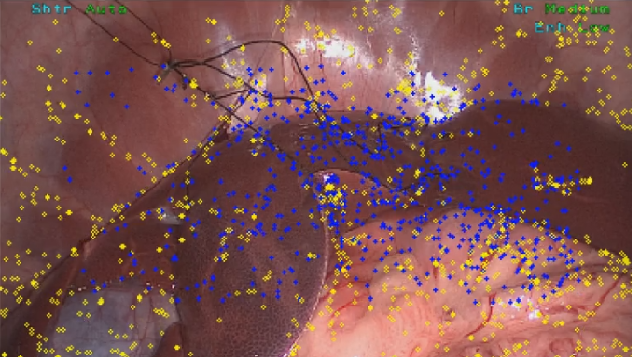}
   }
    \subfigure[]{
    \includegraphics[width=0.25\textwidth]{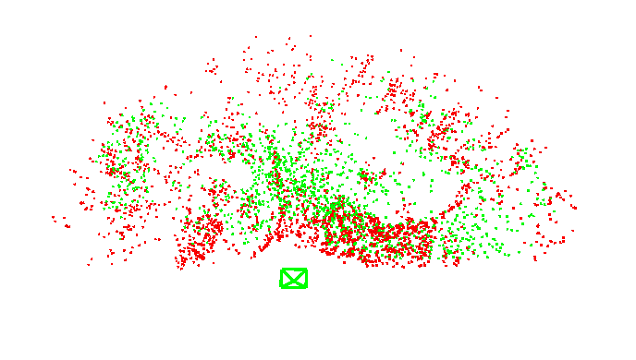}
   }
\caption{Endoscope tracking and reconstructed 3D map during different deformations}
\label{fig:result2}
\end{figure}

Fig. \ref{fig:result2} shows results on different sequences including different deformations and partial scene occlusion due to an instrument. Fig. \ref{fig:result2}(c) shows, from a top view, the endoscope position w.r.t the reconstructed 3D map. The reconstructed 3D map, keyframes and current endoscope position for Fig. \ref{fig:result2}(d) are displayed in Fig. \ref{fig:result2}(e,f). For first row sequence, the size of the reconstructed map were 6750 points, 3263 points for second row sequence and 3740 points for third row sequence.

\vspace{-4mm}
\vspace{-4mm}

It is worth noting that the liver of the second pig (cf. Fig. \ref{fig:result2} second and third rows) was totally texture-less and it was hardly to detect features on its surface, but the system was able to reconstruct many points on liver. The endoscope location was successfully tracked during the interaction with liver in all sequences (cf. Fig. \ref{fig:result2}(b,d,g)). In case of tracking failure due to feature deletion during fast endoscope movements the system was able to relocate the endoscope location once the endoscope had moved and few ORB features were detected. Algorithm 1 is allocated in the  tracking thread, increasing its computation time as shown in Table 1, which reports the average  additional  time due to the  reconstruction and matching of all points.

\begin{figure}
\centering
  \subfigure[]{
    \includegraphics[width=0.22\textwidth]{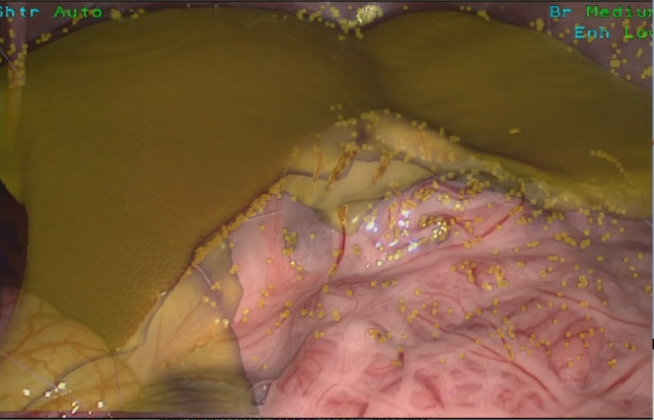}
   }	
  \subfigure[]{
    \includegraphics[width=0.22\textwidth]{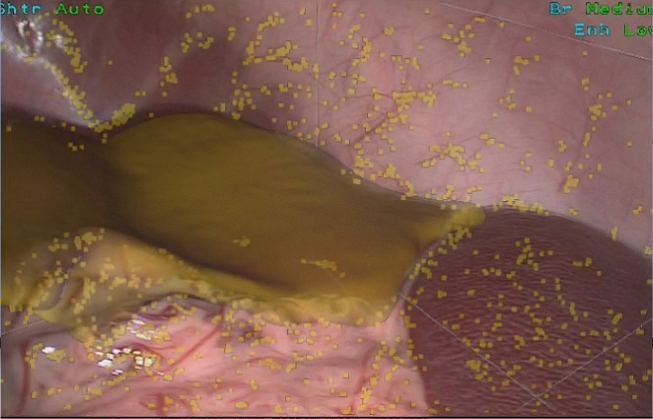}
   }
%    \subfigure[]{
%     \includegraphics[width=0.25\textwidth]{AR3.png}
%    }
   \subfigure[]{
    \includegraphics[width=0.22\textwidth]{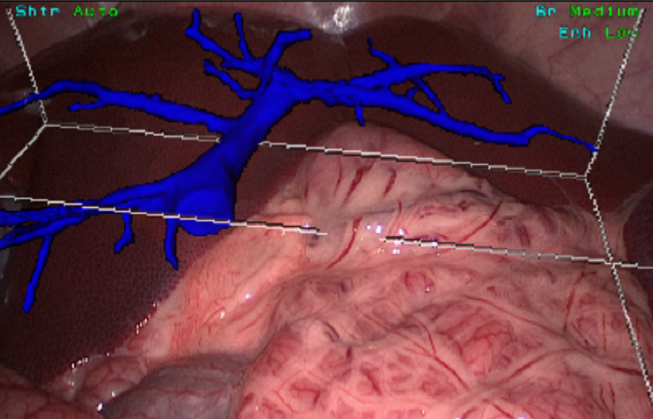}
   }
   \subfigure[]{
    \includegraphics[width=0.22\textwidth]{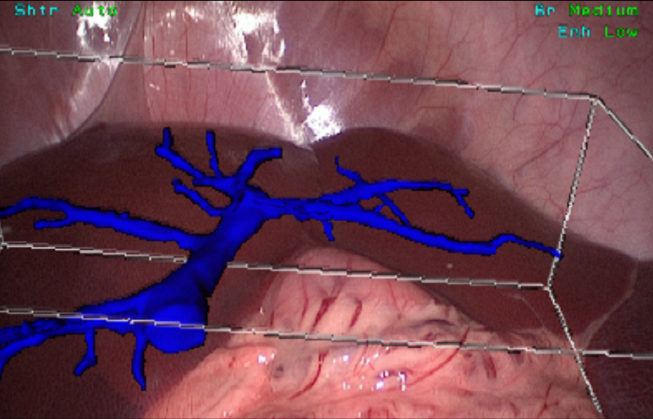}
   }
\caption{Pre-operative data AR overlays. (a-b) liver pre-operative surface segmented and reconstructed from CT images. [c-d] superimposition of liver hepatic veins.}
\label{fig:result3}
\end{figure}

The estimated endoscope location was used to superimpose AR onto one video sequence. The AR insertion was the liver pre-operative surface segmented from CT images in addition to hepatic veins. The pre-operative liver surface and hepatic veins were manually registered in first few frames, and then successfully tracked through out the whole video. Few frames are randomly picked to show the augmented results in Fig. \ref{fig:result3}.
\vspace{-3mm}
\begin{table}
\begin{center}
\caption{Average time (in ms) of different tasks}
\begin{tabu} to \textwidth { | X[c] | X[c] || X[c] | X[c] | X[c] |}
%\begin{tabular}{ |c|c|c|c|c| }
\hline
\multicolumn{2}{|c|}{Mapping} & \multicolumn{3}{|c|}{Tracking} \\
\hline
New points triangulation & ORB triangulation & ORB matching & L-K opt. flow \& cross corr. & Tracking time \\
 \hline
25.3  & 379.2  & 13.3 & 66.2 & 105.2  \\
\hline
\end{tabu}
%\end{tabular}
\end{center}
\end{table}

\vspace{-12mm}
\vspace{-9mm}

\section{Conclusion and Future Work}
In this paper, ORBSLAM system has been re-tuned, proving it as a robust method for monocular endoscope tracking and 3D scene reconstruction from the only input of image stream gathered by the endoscope. Additionally, it is extended to reconstruct a semi-dense map of the scene. The  scene map accuracy has been evaluated against CT ground truth surface and achieving 3-4.1 mm RMSE. The system has also been tested in several in-vivo sequences where displayed a robust performance, even during partial occlusions and severe deformations. In future work, the obtained semi-dense map and the tracked 2D points in the image will be used to estimate the non-rigid organ deformations using shape from template techniques.
\\
\\
\noindent
\textbf{Acknowledgments:} This work is supported by the Direcc{\'i}on General de Investigac{\'i}on Cent{\'i}fica y T{\'e}cnica of Spain under Project RT-SLAM DPI2015-67275-P
\\
\\
\noindent
\textbf{Ethical approval:} All applicable international, national, and/or institutional guidelines for the care and use of animals were followed.

\end{document}